\documentclass[11pt]{article}

\usepackage[preprint]{acl}

\usepackage{times}
\usepackage{latexsym}

\usepackage[T1]{fontenc}
\usepackage{textcomp}

\usepackage[utf8]{inputenc}
\usepackage{amssymb} 
\usepackage{microtype}
\usepackage{amsmath}
\usepackage{booktabs}
\usepackage{array}
\usepackage{enumitem}
\usepackage{tikz}
\usepackage{subcaption}
\usepackage{inconsolata}

\usepackage{graphicx}
\usepackage{amsfonts}
\usepackage{multirow}
\usepackage{xcolor}
\newcommand{\teacher}[1]{\textcolor{gray}{#1}}
\title{MCD: Causal Distillation of Multimodal In-Context Learning in Large Vision-language Models}

\author{
 \textbf{Yanshu Li\textsuperscript{1}},
 \textbf{Jiaqian Li\textsuperscript{1}},
 \textbf{Canran Xiao\textsuperscript{1}},
 \textbf{Xi Xiao\textsuperscript{2}},
 \textbf{Tianyang Wang\textsuperscript{2}},
 \textbf{Yongtai Liu\textsuperscript{3}}\\
 \textsuperscript{1}Brown University,
 \textsuperscript{2}University of Alabama at Birmingham,
 \textsuperscript{3}Hanyang University
\\
 \small{
   \textbf{Correspondence:} \href{mailto:yanshu\_li1@brown.edu}{yanshu\_li1@brown.edu}
 }
}

\begin{document}
\maketitle
\begin{abstract}
Large vision-language models (LVLMs) exhibit strong multimodal in-context
learning (ICL) capabilities, yet this ability degrades substantially as model
size decreases. Knowledge distillation offers a natural way to bridge this gap, but existing methods primarily align output distributions or hidden representations directly. Such alignment teaches the student what the teacher predicts without revealing which evidence in the complex context causally supports that prediction. Consequently, a student can imitate the
teacher's answer while continuing to rely on language priors, prompt structure, or other spurious cues. To address this limitation, we introduce
\emph{Multimodal Causal Distillation} (MCD), a distillation framework that
transfers how a strong teacher uses multimodal evidence during ICL. MCD uses structure-preserving token interventions to identify and verify causal evidence, then transfers how the teacher responds when that evidence is retained or removed. This design connects distillation to the causal patterns by which the model uses contextual evidence during multimodal ICL. Experiments across three LVLM families and seven benchmarks show that MCD improves student performance by 7.23 points on average and outperforms vanilla distillation by 4.68 points, while further analyses confirm the generalizability of these gains.
\end{abstract}

\section{Introduction}
Large vision-language models (LVLMs) have demonstrated remarkable capabilities
in multimodal understanding and reasoning tasks \cite{vlm}. An increasingly salient capability is multimodal in-context learning (ICL), in
which a model answers a query using a small set of interleaved image-text
demonstrations provided in the prompt \cite{mmicl}. It provides a
flexible way of adapting a general-purpose LVLM to diverse tasks at
inference time. Its effectiveness, however, depends strongly on model scale \cite{makes}. Larger LVLMs can integrate the query with relevant evidence across multiple demonstrations, whereas smaller models are more susceptible to spurious cues such as language priors and prompt formatting \cite{small2}. Prior work has sought to improve multimodal ICL through prompt configuration, but its effectiveness remains constrained by the model’s intrinsic capabilities \cite{how, makes}.

Thus, strong-to-weak knowledge distillation offers a better approach to narrowing this capability gap by transferring the behavior of a powerful teacher to a compact student \cite{kd}. Most existing distillation methods supervise the student by matching the teacher's output distribution. However, in multimodal ICL with complex context, this pointwise constraint leaves a critical ambiguity unresolved \cite{mmd}. The same answer may arise from integrating task-relevant contextual evidence or exploiting superficial correlations, such as textual biases, demonstration order, and output-format regularities. Consequently, a weak student may match the teacher on training prompts without learning how the teacher uses provided context. Another paradigm distills intermediate attention patterns to provide finer-grained supervision~\cite{compo}, but incurs substantial costs on multi-image prompts and introduces additional training instability in multimodal ICL.

To develop a distillation method tailored to multimodal ICL, we study this problem from a causal perspective. Given a query and in-context demonstrations, the model must combine the task specified by the query with mechanism-relevant evidence from the demonstrations. This motivates structure-preserving interventions that modify semantic content and measure the resulting output changes. Causal evidence should preserve the teacher’s prediction when retained and substantially alter it when removed, providing supervision unavailable from the original prompt alone. However, identifying such evidence without annotations or exhaustive token removal is challenging \cite{cama}, particularly for multi-image prompts, while arbitrary replacements may confound semantic effects with structural corruption.

To this end, we propose \emph{Multimodal Causal Distillation} (MCD), which transfers how a strong teacher causally uses multimodal evidence during ICL. MCD first separates prompt structure from content-bearing text and visual tokens, then applies attribute-matched replacements that preserve each token’s functional or spatial position while altering its semantics. It uses teacher gradients to identify query and demonstration evidence supporting the teacher’s answer and verifies this evidence through complementary interventions that retain or remove the selected tokens. Valid evidence should preserve the teacher’s prediction when retained and induce a larger output change when removed. MCD transfers both behaviors to the student, thereby teaching evidence sufficiency and causal dependence without requiring mechanism annotations or attention alignment. Extensive experiments with three LVLM families and seven benchmarks demonstrate MCD’s superior performance, validating its effectiveness and generality. Our main contributions can be summarized as follows:
\begin{itemize}
    \item We propose Multimodal Causal Distillation (MCD), the first distillation framework that formulates multimodal ICL from a causal perspective and transfers how a strong teacher causally uses multimodal evidence, moving beyond output-only alignment.
    \item MCD combines structure-preserving token interventions, gradient-based evidence discovery, and retain-remove verification to effectively transfer both evidence sufficiency and causal dependence to the student.
    \item Extensive experiments across three LVLM families and seven benchmarks show that MCD improves student performance by 7.23 on average and outperforms vanilla distillation by 4.68, confirming the generality of causal supervision for multimodal ICL.
\end{itemize}
\section{Related Works}
\label{sec2}
\paragraph{Multimodal in-context learning (ICL).} Recent LVLMs have evolved into general-purpose systems capable of complex multimodal tasks~\cite{survey}. One key capability is multimodal ICL, which enables models to infer visual grounding rules, output formats, and input-output mappings from a few examples without parameter updates~\cite{rela1,rela2}. However, multimodal ICL remains unstable, especially in smaller models that often rely on textual cues or superficial structural patterns rather than the multimodal context~\cite{rela3,micl1,micl2,small2}. Existing approaches primarily optimize prompt configuration and therefore provide limited improvements to the model’s underlying reasoning capabilities~\cite{rela5,rela6}.
\paragraph{Knowledge distillation.} Knowledge distillation transfers knowledge from strong teachers to compact students \cite{kd,kd2,kd3} and has recently been extended from LLMs to LVLMs~\cite{rela8}. Existing methods often model fine-grained interactions between visual and textual tokens~\cite{rela7}. Align-KD aligns cross-modal attention~\cite{alignkd}, and LLaVA-KD introduces relational distillation~\cite{llavakd}. Meanwhile, CompoDistill attempts to leverage LVLM’s internal attention distributions~\cite{compo}. Align-TI~\cite{alignti} combines both strategies to capture multiple forms of token interactions. However, their reliance on the teacher’s attention maps may not faithfully capture the causal dependencies. Among LLM distillation methods, LeaF~\cite{leaf} moves beyond attention by using interventions to expose teacher–student gradient differences and reveal token interactions. It is effective on long-context text tasks, but its reliance on pruning-based interventions limits its applicability to multimodal ICL.

\section{Method}
\label{sec:method}

\subsection{Overview}
\label{sec:overview}
\begin{figure*}[t]
    \centering
    \includegraphics[width=\textwidth]{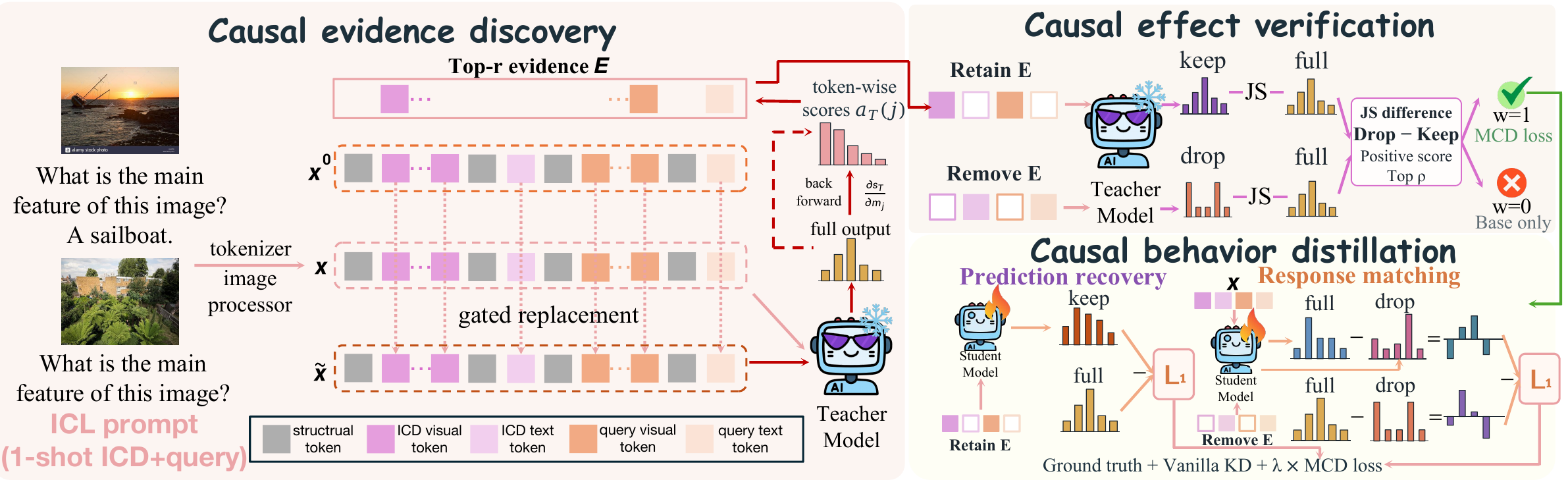}
    \caption{Overview of the proposed Multimodal Causal Distillation (MCD) framework.}
    \label{fig:pipeline}
    \vspace{-15pt}
\end{figure*}
We propose \emph{Multimodal Causal Distillation} (MCD), illustrated in Fig.~\ref{fig:pipeline}, to transfer how a strong teacher uses multimodal ICL evidence to a smaller student. We first motivate MCD from a causal perspective in Section~\ref{sec:3.2}. We then discover mechanism-bearing evidence through structure-preserving interventions in Section~\ref{sec:3.3} and verify it using retain-remove tests in Section~\ref{sec:effect_estimation}. Finally, we transfer evidence sufficiency and causal dependence in Section~\ref{sec:causal_distillation} and integrate them with supervised learning and output distillation in Section~\ref{sec:overall_objective}.

\subsection{Preliminaries and Motivation}
\label{sec:3.2}

\subsubsection{Vanilla Vision-language Distillation}
\label{sec:vl_distillation}

LVLMs considered in this work follow a widely adopted architecture consisting
of a vision encoder, a projector, and an LLM backbone. As discussed in
Section~\ref{sec2}, such models are susceptible to modality bias and sensitive
to prompt formatting at smaller scales, leading to substantial deficiencies
in multimodal reasoning. These deficiencies become more pronounced in
multimodal ICL, which requires complex cross-modal interactions. Formally, the
input $\boldsymbol{x}$ of multimodal ICL comprises $n$ in-context
demonstrations (ICDs) and a query sample,
\begin{equation}
    \boldsymbol{x}
    =[\mathcal{C};q]=
    \left[
        \{(I_i,T_i)\}_{i=1}^{n};
        (\hat I,\hat T)
    \right],
    \label{eq:mmicl_input}
\end{equation}
where each ICD contains an image $I_i$ and a text segment $T_i$ composed of a
question and its answer label. The query contains an image
$\hat I$ and an unlabeled question $\hat T$. At inference time, the LVLM is expected to use the task-relevant external knowledge, output format, and input-output mappings conveyed by the ICDs~\cite{taco} to infer
the answer to the query sample. We refer to this unobserved
information as the \emph{latent task mechanism}.

Given the substantial gap in multimodal ICL capability observed between
large-scale LVLMs and their smaller counterparts from the same
family~\cite{true,vlicl}, distillation provides a natural way to improve
small-scale LVLMs. Let the large-scale teacher define an output distribution
$P_T$, and let $P_S^\theta$ denote the distribution of a student parameterized
by $\theta$. For a training sample $(\boldsymbol{x},\boldsymbol{y})$ drawn from
$\mathcal{D}$, define
$P_{T,k}=P_T(\cdot\mid\boldsymbol{x},\boldsymbol{y}_{<k})$ and
$P_{S,k}^{\theta}
=P_S^\theta(\cdot\mid\boldsymbol{x},\boldsymbol{y}_{<k})$.
Vanilla output distillation minimizes
\begin{equation}
    \mathcal{L}_{\mathrm{dis}}(\theta)
    =
    \mathbb{E}_{(\boldsymbol{x},\boldsymbol{y})\sim\mathcal{D}}
    \left[
        \frac{1}{L}
        \sum_{k=1}^{L}
        D_{\mathrm{KL}}
        \left(P_{T,k}\middle\|P_{S,k}^{\theta}\right)
    \right],
    \label{eq:standard}
\end{equation}
where $\boldsymbol{y}_{<k}$ denotes the ground-truth prefix before the
$k$-th decoding step and $L$ is the answer length. It enables the student to
imitate the output behavior of a stronger teacher. However, alignment only at
the output level provides insufficient supervision for the student to learn
how to exploit complex cross-modal semantics, thereby limiting its
generalization. To address this limitation, we first conduct a causal analysis
of multimodal ICL.

\subsubsection{A Causal View of Multimodal ICL}
\label{sec:causal_view}

\begin{figure}[t]
    \centering
    \includegraphics[width=\columnwidth]{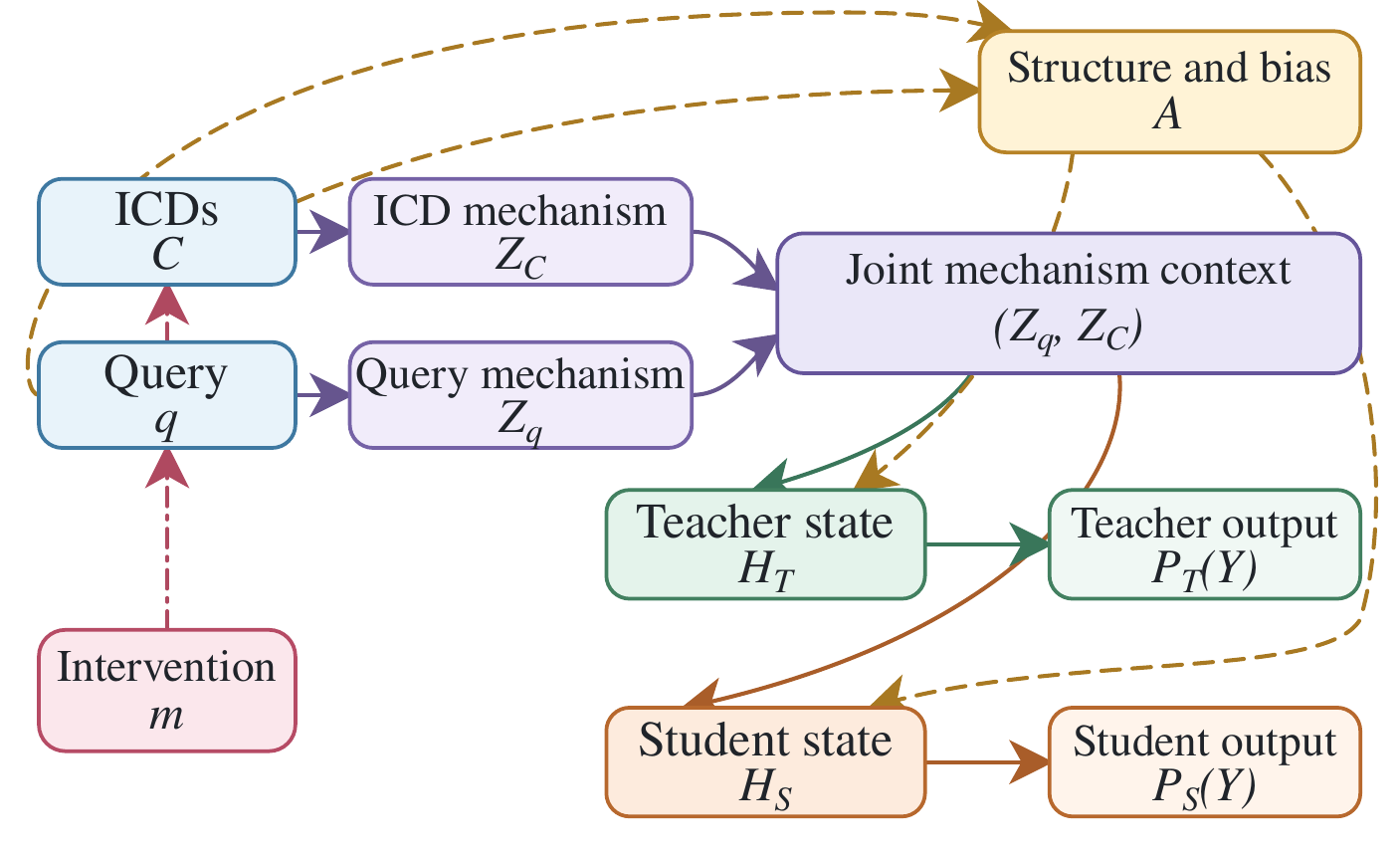}
    \caption{
        Causal graph of multimodal ICL inference. The query and ICDs provide
        complementary mechanism evidence and jointly shape the model state.
    }
    \label{fig:causal_graph}
    \vspace{-15pt}
\end{figure}

Given a prompt $[\mathcal{C};q]$, an LVLM must interpret the mechanism
indicated by $q$ using the mechanism evidence conveyed by $\mathcal{C}$.
This inference process entails a causal dependence: the mechanisms conveyed
by the query and ICDs jointly shape the model state and its prediction, while
prompt structure and modality bias can provide competing paths to the output.
To make these dependencies explicit, we formulate multimodal ICL inference
with the causal graph in Fig.~\ref{fig:causal_graph}. For a model
$M\in\{T,S\}$, the inference relations are
\begin{equation}
    \begin{aligned}
        H_M
        &=
        f_M(Z_q,Z_{\mathcal{C}},\boldsymbol{A}),\\
        P_M(Y\mid\boldsymbol{x})
        &=
        p_M(Y\mid H_M).
    \end{aligned}
    \label{eq:causal_scm}
\end{equation}
Here $Z_q$ and $Z_{\mathcal{C}}$ summarize the latent task mechanism indicated
by the query and the evidence conveyed by the ICDs. Their joint effect on
$H_M$ represents the model's mechanism-conditioned use of the prompt, while
$\boldsymbol{A}$ captures structural and modality cues induced by prompt
construction.

This graph allows us to distinguish the causal pathways underlying multimodal
ICL. An intervention $\boldsymbol{m}$ replaces selected query or ICD content
upstream of $Z_q$ and $Z_{\mathcal{C}}$ while preserving
$\boldsymbol{A}$. The resulting output change provides a measure of how strongly the model’s prediction depends on the intervened content. In contrast, Eq.~\eqref{eq:standard} aligns the teacher and student only
on the observed prompt. A weak student may match the teacher's answer while
still relying on spurious biases. Thus, effective distillation should transfer
the teacher's causal response to mechanism-bearing evidence in addition to
its observed output distribution. To this end, we propose
MCD, which employs interventions
derived from the causal relations.

\subsection{Causal Evidence Discovery}
\label{sec:3.3}

First, we intervene on the token-level semantics of $q$ and $\mathcal{C}$ in
Fig.~\ref{fig:causal_graph}. We partition the input prompt into structural tokens and candidate tokens $\mathcal{U}(\boldsymbol{x})=\{u_j\}_{j=1}^{N}$. Structural tokens comprise
the template tokens required to execute the prompt and remain unchanged under
all interventions. Candidate tokens comprise the content-bearing text and
projected visual tokens from both the ICDs and the query. Thus, $N$ is
determined by the prompt rather than set as a hyperparameter. For each candidate token $u_j$, we record attributes used to select its replacement, as detailed in Appendix~\ref{app:replacement}. 

During preprocessing, each prompt $\boldsymbol{x}$ is randomly paired with
another training example $\boldsymbol{x}^0$ that has the same prompt layout,
and this pairing is reused across all training epochs. For each $u_j$, we
select the token $u_j^0$ from $\boldsymbol{x}^0$ with the corresponding
attributes. Let $e_{M,j}$ and $b_{M,j}$ denote the model-specific embeddings
of $u_j$ and $u_j^0$, respectively. A scalar gate defines
\begin{equation}
    \widetilde e_{M,j}(m_j)
    =
    b_{M,j}
    +
    m_j(e_{M,j}-b_{M,j}),
    M\in\{T,S\}.
    \label{eq:content_gate}
\end{equation}
Setting $m_j=1$ retains the original token, while $m_j=0$ replaces it. The
teacher and student use the same original and replacement tokens but encode
them with their own input modules. Thus, the intervention changes the token content while preserving the structural and positional properties specified by its attributes, thereby avoiding the introduction of additional biases. In terms of Fig.~\ref{fig:causal_graph}, it changes
the inputs to $Z_q$ and $Z_{\mathcal{C}}$ while keeping the structural factors represented by $\boldsymbol{A}$.

We next identify which candidate tokens causally contribute to the teacher's predictions in multimodal ICL. The teacher first receives the original prompt and generates a response $\boldsymbol{y}_T=(y_{T,1},\ldots,y_{T,L_T})$, where $L_T$ denotes the response length. We keep this response and its prefixes fixed when
evaluating all interventions and define the teacher score as
\begin{equation}
   \small
    s_T(\boldsymbol{x};\boldsymbol{m})
    =
    \frac{1}{L_T}
    \sum_{k=1}^{L_T}
    \log
    P_T\!\left(
        y_{T,k}
        \mid
        \boldsymbol{x},
        \boldsymbol{y}_{T,<k};
        \boldsymbol{m}
    \right).
    \label{eq:teacher_score}
\end{equation}

To score all candidate tokens with one backward pass, we sample
$\alpha\sim\mathcal{U}(0,1)$ and set every gate to $\alpha$. The importance of
$u_j$ is
\begin{equation}
\begin{aligned}
    a_T(j)
    &=
    \left.
    \frac{\partial s_T(\boldsymbol{x};\boldsymbol{m})}
         {\partial m_j}
    \right|_{\boldsymbol{m}=\alpha\boldsymbol{1}}
    \\
    &=
    \bigl(e_{T,j}-b_{T,j}\bigr)^\top
    \nabla_{\widetilde e_{T,j}}
    s_T\bigl(\boldsymbol{x};\alpha\boldsymbol{1}\bigr).
\end{aligned}
\end{equation}
A larger $a_T(j)$ indicates a stronger local increase in the teacher score when $u_j$ moves from its replacement toward its original content at the sampled interpolation point. We rank the candidate tokens by this score and select
\begin{equation}
\begin{aligned}
    E
    &=
    \operatorname{Top}_{m_E}
    \left\{j:a_T(j)>0\right\}_{j=1}^{N},
    \\
    m_E
    &=
    \max\left\{1,\left\lfloor rN\right\rfloor\right\},
0<r<1.
\end{aligned}
\label{eq:evidence_selection}
\end{equation}
where $r$ is the fixed fraction of candidate tokens retained as evidence. Examples with $E=\varnothing$ are not eligible for causal-effect verification and are assigned $w(\boldsymbol{x})=0$. The
causal role of $E$ is verified next through discrete interventions.

\subsection{Causal Effect Verification}
\label{sec:effect_estimation}

The gradient score indicates which tokens support the teacher locally, but it
does not by itself establish that these tokens determine the teacher's
prediction. We therefore evaluate two complementary prompt variants. The
\emph{retain-evidence} variant keeps the tokens in $E$ and replaces every
other candidate token. The \emph{remove-evidence} variant replaces the tokens
in $E$ and keeps every other candidate token. Their gate vectors are
\begin{equation}
    \small
    m_j^{K}
    =
    \mathbf{1}[j\in E],
    m_j^{D}
    =
    1-\mathbf{1}[j\in E],
    m_j^{F}=1,
\end{equation}
where $F$ denotes the original prompt, $K$ the retain-evidence prompt, and
$D$ the remove-evidence prompt. Structural tokens remain unchanged in all
three prompt settings.

For $v\in\{F,K,D\}$, let
\begin{equation}
    P_{T,k}^{v}
    =
    P_T\!\left(
        \cdot
        \mid
        \boldsymbol{x},
        \boldsymbol{y}_{T,<k};
        \boldsymbol{m}^{v}
    \right).
\end{equation}
We measure whether $E$ is sufficient and necessary for the teacher using Jensen–Shannon divergence:
\begin{align}
    d_T^{\mathrm{keep}}
    &=
    \frac{1}{L_T}
    \sum_{k=1}^{L_T}
    D_{\mathrm{JS}}
    \left(
        P_{T,k}^{F}
        \middle\|
        P_{T,k}^{K}
    \right),
    \label{eq:keep_effect}\\
    d_T^{\mathrm{drop}}
    &=
    \frac{1}{L_T}
    \sum_{k=1}^{L_T}
    D_{\mathrm{JS}}
    \left(
        P_{T,k}^{F}
        \middle\|
        P_{T,k}^{D}
    \right).
    \label{eq:drop_effect}
\end{align}
A small $d_T^{\mathrm{keep}}$ means that $E$ alone preserves the teacher's
prediction, while a large $d_T^{\mathrm{drop}}$ means that removing $E$
changes it. Because both quantities are measured using the same bounded divergence, we combine them directly as follows:
\begin{equation}
    c_T
    =
    d_T^{\mathrm{drop}}
    -
    d_T^{\mathrm{keep}}.
    \label{eq:causal_reliability}
\end{equation}
We retain the top fraction $\rho$ of training examples with $c_T>0$ and assign them $w(\boldsymbol{x})=1$. All remaining examples receive $w(\boldsymbol{x})=0$ and use only the standard supervised terms. Thus, causal supervision is applied only when retaining the selected tokens reproduces the full teacher response more faithfully than removing them. These verified tokens provide an operational link to the latent mechanism because they carry the query and ICD semantics that the teacher actually uses, as evidenced by the causal changes in its prediction when these tokens are intervened upon.

\subsection{Causal Behavior Distillation}
\label{sec:causal_distillation}

For each accepted example, MCD teaches the student two behaviors established
by the verification step. First, the student should recover the teacher's
full-prompt prediction when only $E$ is retained. Second, removing $E$ should
change the student prediction in the same way that it changes the teacher
prediction. We evaluate both models on the fixed teacher-generated prefixes.

To reduce offline storage, we retain the union of the teacher's top
$K_{\mathrm{cache}}$ output tokens under $F$ and $D$ at each decoding step and
aggregate the remaining probability mass into a single tail entry. We apply
the same deterministic aggregation to the student distributions. Let
$\bar P_{M,k}^{v}$ denote the resulting distribution for model
$M\in\{T,S\}$ and prompt variant $v\in\{F,K,D\}$. This representation exactly preserves the total probability mass assigned to the cached support and the aggregated tail, providing a memory-efficient approximation to the corresponding full-vocabulary losses. Conventional distillation probabilities are stored in the same form using a separate support constructed at the ground-truth prefixes.

We define the change caused by removing $E$:
\begin{equation}
    \boldsymbol{\Delta}_{M,k}
    =
    \bar P_{M,k}^{F}
    -
    \bar P_{M,k}^{D}.
\end{equation}
Using this causal change, we define two causal distillation terms as follows:
\begin{align}
    \ell_{\mathrm{keep},k}
    &=
    \frac{1}{2}
    \left\|
        \bar P_{T,k}^{F}
        -
        \bar P_{S,k}^{K}
    \right\|_1,
    \label{eq:student_keep_loss}\\
    \ell_{\mathrm{effect},k}
    &=
    \frac{1}{4}
    \left\|
        \boldsymbol{\Delta}_{T,k}
        -
        \boldsymbol{\Delta}_{S,k}
    \right\|_1.
    \label{eq:student_effect_loss}
\end{align}
The first term transfers evidence sufficiency. The second transfers the
teacher's causal response to removing that evidence. Both terms lie in
$[0,1]$. We combine them to obtain the core MCD loss:
\begin{equation}
    \mathcal{L}_{\mathrm{MCD}}
    =
    \frac{w(\boldsymbol{x})}{2L_T}
    \sum_{k=1}^{L_T}
    \left(
        \ell_{\mathrm{keep},k}
        +
        \ell_{\mathrm{effect},k}
    \right).
    \label{eq:full_mcd_loss}
\end{equation}

Computing both terms for every accepted example would require three student
evaluations on teacher-generated prefixes: one each for $F$, $K$, and $D$.
We reduce this cost by sampling
$z\sim\operatorname{Bernoulli}(1/2)$ and optimizing
\begin{equation}
    \widehat{\mathcal{L}}_{\mathrm{MCD}}
    =
    \frac{w(\boldsymbol{x})}{L_T}
    \sum_{k=1}^{L_T}
    \left[
        z\,\ell_{\mathrm{keep},k}
        +
        (1-z)\,\ell_{\mathrm{effect},k}
    \right].
    \label{eq:sampled_mcd_loss}
\end{equation}
When $z=1$, the student is evaluated only on $K$. When $z=0$, it is evaluated
on $F$ and $D$. Therefore,
\begin{equation}
    \mathbb{E}_{z}
    \left[
        \widehat{\mathcal{L}}_{\mathrm{MCD}}
    \right]
    =
    \mathcal{L}_{\mathrm{MCD}},
    \label{eq:unbiased_estimator}
\end{equation}
and each accepted example needs an average of $1.5$ student evaluations on
teacher-generated prefixes.

\subsection{Training Objective and Efficiency}
\label{sec:overall_objective}
The overall training objective combines supervised learning, vanilla
distillation, and causal distillation:
\begin{equation}
    \mathcal{L}
    =
    \mathcal{L}_{\mathrm{sup}}
    +
    \mathcal{L}_{\mathrm{dis}}
    +
    \lambda
    \widehat{\mathcal{L}}_{\mathrm{MCD}},
    \label{eq:overall_loss}
\end{equation}
where $\mathcal{L}_{\mathrm{sup}}$ is the ground-truth cross-entropy and
$\mathcal{L}_{\mathrm{dis}}$ is defined in
Eq.~\eqref{eq:standard}. The loss coefficient $\lambda$ controls the contribution of MCD to the overall training objective. After distillation, the student can reason more effectively over token-level evidence during multimodal ICL inference, improving its ability to utilize complex cross-modal context.

All teacher generations, verification scores, and output
distributions are computed once and cached before student training. MCD operates only on input embeddings and output distributions and requires no alignment of attention matrices. Thus, it can be applied to diverse model architectures, such as backbones with linear attention.

\section{Experiments}
\subsection{Setup}
\paragraph{Training data.}To construct a high-quality training set for MCD, we first include all samples from three benchmarks specifically designed to evaluate multimodal ICL: VL-ICL \cite{vlicl}, TrueMICL \cite{true}, and SMMILE \cite{smmile}. In their few-shot instances, LVLMs cannot answer the query without first extracting sufficient task information from the corresponding ICDs. We then use TACO~\cite{taco} to construct additional multimodal few-shot prompts from HatefulMemes \cite{hateful}, MME-RealWorld \cite{mme}, BlindTest \cite{blind}, VisuLogic \cite{visu}, and GQA \cite{gqa}, further expanding the training set to 90K prompts. Finally, we remove duplicate prompts and retain only those for which the teacher correctly answers the query given the original prompt, resulting in 60K training examples. Among these prompts, 80\% contain four ICDs, 10\% contain eight ICDs, and 5\% each contain one and two ICDs.
\paragraph{Models and benchmarks.}We evaluate MCD across three LVLM families: LLaVA-OneVision~\cite{ov}, Qwen3-VL~\cite{3vl}, and Qwen3.5~\cite{3.5}. Within each family, MCD distills multimodal ICL capabilities from a larger teacher into a smaller student using official few-shot templates. We evaluate on benchmarks sufficiently out of distribution from the training data so that performance gains reflect causal ICL behavior instead of knowledge transfer alone. We evaluate on VQAv2~\cite{vqa}, VizWiz~\cite{vizwiz}, MMStar~\cite{mmstar}, MathVision~\cite{mathvision}, MDK12~\cite{mdk12}, MMIQ~\cite{mmiq}, and LogicVista~\cite{logicvista}. VQAv2 and VizWiz are widely used to evaluate multimodal ICL, while the remaining five benchmarks are frequently adopted in recent evaluations of state-of-the-art LVLMs. For benchmarks with predefined training and test splits, we use the training split as the ICD pool and the test split as the query set. For benchmarks without predefined splits, we partition the samples into ICD and query sets at a ratio of 6:4. We use TACO to configure the ICD sequences, and all main experiments are conducted in the four-shot setting.
\paragraph{Baselines.} Following standard evaluation protocols for knowledge distillation, we report the performance of the teacher, the student before and after MCD, and the student trained with vanilla distillation. Since no prior distillation method specifically targets multimodal ICL, we compare MCD with recent methods for general LVLM distillation that can be directly applied to multi-image settings, including LLaVA-KD \cite{llavakd}, CompoDistill \cite{compo}, and Align-TI \cite{alignti}. For Qwen3.5, the attention-based objectives in Align-TI are applied only to full-attention layers, while CompoDistill is not applied to this model.
\paragraph{Implementation details.} We train each model for three epochs using a customized Hugging Face Trainer. We freeze the vision encoder and perform full-parameter supervised fine-tuning on all remaining modules. We set the batch size to 32, $\lambda$ to 1.0, $r$ to 0.25, $\rho$ to 0.5, and $K_{\text{cache}}$ to 128. During evaluation, the teacher and student within each model family use greedy decoding with identical inference settings. Given the instability of multimodal ICL, we report the average results over three different random seeds in the main experiments.
\begin{table*}[t]
    \centering

    \resizebox{\textwidth}{!}{
    \begin{tabular}{clcccccccc}
        \toprule
        Model Family
        & Model / Method
        & VQAv2
        & VizWiz
        & MMStar
        & MathVision
        & MDK12
        & MMIQ
        & LogicVista
        & Avg. \\
        \midrule

        \multirow{7}{*}{\shortstack[l]{LLaVA-OneVision}}
        & Teacher (72B)
        & \teacher{83.61} & \teacher{71.87} & \teacher{61.05}
        & \teacher{27.32} & \teacher{46.57} & \teacher{28.03}
        & \teacher{32.14} & \teacher{50.08} \\
        & Student (7B)
        & 78.00 & 63.32 & 51.17 & 17.52 & 38.78 & 23.21 & 25.74 & 42.53 \\
        & +Vanilla KD
        & 81.29 & 67.42 & 52.38 & 18.46 & 40.17 & 24.17 & 27.53 & 44.49 \\
        & +LLaVA-KD
        & 82.61 & 69.23 & 54.73 & 18.77 & 41.33 & 24.14 & 28.95 & 45.68 \\
        & +CompoDistill
        & 81.57 & 69.45 & 53.19 & 17.95 & 40.63 & 23.81 & 28.42 & 45.00 \\
        & +Align-TI
        & 82.74 & 69.89 & 55.26 & 20.47 & 41.73 & 26.42 & 28.86 & 46.48 \\
        & +\textbf{MCD (Ours)}
        & \textbf{82.95} & \textbf{70.17} & \textbf{57.17}
        & \textbf{22.67} & \textbf{44.82} & \textbf{27.38}
        & \textbf{30.46} & \textbf{47.95} \\
        \midrule

        \multirow{7}{*}{\shortstack[l]{Qwen3-VL}}
        & Teacher (32B)
        & \teacher{87.24} & \teacher{77.78} & \teacher{76.62}
        & \teacher{63.28} & \teacher{53.81} & \teacher{37.86}
        & \teacher{62.48} & \teacher{65.58}\\
        & Student (2B)
        & 81.47 & 70.92 & 58.90 & 39.14 & 40.75 & 29.36 & 39.63 & 51.45 \\
        & +Vanilla KD
        & 83.72 & 75.28 & 63.27 & 39.92 & 42.07 & 30.46 & 41.10 & 53.69 \\
        & +LLaVA-KD
        & 83.96 & 75.61 & 65.45 & 42.17 & 45.29 & 31.28 & 43.34 & 55.30 \\
        & +CompoDistill
        & 83.56 & 75.42 & 63.37 & 41.04 & 45.60 & 30.39 & 41.31 & 54.38 \\
        & +Align-TI
        & 84.28 & 76.27 & 65.64 & 43.50 & 47.43 & 33.62 & 43.71 & 56.35 \\
        & +\textbf{MCD (Ours)}
        & \textbf{85.31} & \textbf{77.04} & \textbf{67.26}
        & \textbf{46.27} & \textbf{50.32} & \textbf{36.24}
        & \textbf{45.09} & \textbf{58.22} \\
        \midrule

        \multirow{6}{*}{\shortstack[c]{Qwen3.5\\(non-thinking)}}
        & Teacher (27B)
        & \teacher{92.75} & \teacher{81.63} & \teacher{84.12}
        & \teacher{78.53} & \teacher{63.25} & \teacher{49.29}
        & \teacher{71.37} & \teacher{74.42}\\
        & Student (2B)
        & 84.62 & 75.27 & 68.35 & 42.70 & 49.23 & 30.61 & 47.38 & 56.88 \\
        & +Vanilla KD
        & 85.46 & 76.35 & 71.50 & 49.42 & 53.28 & 33.42 & 52.83 & 60.32 \\
        & +LLaVA-KD
        & 87.79 & 78.21 & 74.29 & 55.26 & 55.62 & 36.71 & 54.92 & 63.26 \\
        & +Align-TI
        & 88.74 & 78.00 & 76.01 & 57.35 & 56.97 & 37.25 & 55.25 & 64.22 \\
        & +\textbf{MCD (Ours)}
        & \textbf{89.27} & \textbf{79.24} & \textbf{78.75}
        & \textbf{58.47} & \textbf{58.53} & \textbf{40.17}
        & \textbf{60.24} & \textbf{66.38} \\
        \bottomrule
    \end{tabular}
    }
        \caption{Four-shot performance of different models on seven multimodal benchmarks. \textbf{Bold} denotes the best student result within each model family. Avg. denotes the average score across all seven benchmarks.}
    \label{tab:main_results}
\end{table*}

\subsection{Main results}
As shown in Table~\ref{tab:main_results}, MCD achieves the best student performance on every benchmark, with average scores of 47.95, 58.22, and 66.38 for LLaVA-OneVision, Qwen3-VL, and Qwen3.5, respectively. These results improve upon the original students by 5.42, 6.77, and 9.50 points. Across the three families, MCD also outperforms Vanilla KD by 3.46 to 6.06 points and the strongest specialized baseline, Align-TI, by 1.47 to 2.16 points. Its advantage over Align-TI is particularly pronounced on challenging benchmarks with larger gaps between teacher and student performance, averaging 2.29 points compared with 0.68 points on the remaining benchmarks. This pattern suggests that transferring the teacher’s causal dependence on mechanism-bearing evidence becomes increasingly valuable as contextual reasoning grows more complex. Overall, MCD delivers larger gains on reasoning-intensive tasks while maintaining strong performance on general visual question answering. Its improvements across all three model families further demonstrate its applicability to LVLMs with different parameter scales and attention architectures.
\subsection{Ablation Study}
\label{sec:ablation}
We conduct ablation studies and analyses on Qwen3.5 and Qwen3-VL and report results averaged across the two model families.

\begin{figure}[t]
    \centering
    \includegraphics[width=\columnwidth]{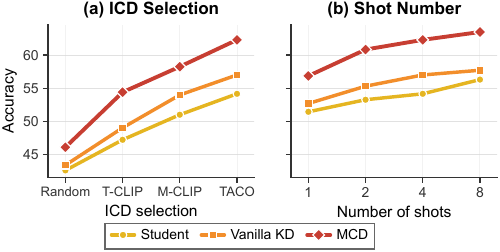}
    \caption{Performance under (a) different ICD selection strategies and (b) numbers of ICDs.}
    \label{fig:shot_retrieval}
\end{figure}

\paragraph{Robustness across ICL configurations.}
We first examine whether the improvement of MCD transfers across different
ICL prompt configurations. For each query, we ablate both the number of ICDs and the ICD retrieval strategy. To vary the number of ICDs, we additionally construct 1, 2, and 8-shot prompts. To vary retrieval quality, we use random sampling, text-based CLIP similarity (T-CLIP), and joint image-text CLIP similarity (M-CLIP), which generally produce prompts of increasing quality \cite{how}. Fig.~\ref{fig:shot_retrieval}
reports the average scores over the seven benchmarks for the original student, Vanilla KD, and MCD. MCD maintains substantial gains over both baselines across diverse configurations, with its performance further improving as prompt quality increases. The comparisons jointly evaluate robustness to the quantity and quality of in-context evidence, demonstrating the broad applicability of MCD across diverse user requirements.

\paragraph{Contributions of causal supervision.}
We next isolate the two causal objectives and the verification procedure. We compare full MCD with its variants that remove either $\ell_{\mathrm{keep}}$ or $\ell_{\mathrm{effect}}$ or disable causal-effect verification while retaining both objectives. As shown in Table~\ref{tab:causal_supervision_ablation}, MCD scores 66.34 on average, dropping to 64.89 without $\ell_{\mathrm{keep}}$, 64.36 without $\ell_{\mathrm{effect}}$, and 63.47 without verification. The complementary drops indicate that recovering the teacher prediction from retained evidence and matching its response to evidence removal capture different aspects of causal behavior, with the latter contributing more strongly to reasoning-intensive tasks. The larger degradation without verification further shows that causal supervision is most effective when the selected evidence induces a reliable teacher response, motivating MCD to verify interventions before transferring them to the student.

\paragraph{Design of causal evidence discovery.}
We then ablate teacher-gradient ranking and attribute-matched replacement in Table~\ref{tab:evidence_discovery_ablation}. Replacing gradient ranking with random or attention-based selection reduces the average score from 66.34 to 62.26 and 64.59, respectively, while unrestricted replacement yields 63.85. The substantial degradation under random selection confirms that the effectiveness of causal supervision depends on identifying content that meaningfully affects the teacher prediction, while the remaining gap of attention-based selection indicates that teacher-response gradients provide a more direct signal of this dependence. The decline caused by unrestricted replacement further shows the need to preserve the functional and spatial roles of intervened tokens, allowing the measured output change to be attributed to semantic evidence rather than unintended prompt corruption.

\begin{table}[t]
    \centering
    \resizebox{\columnwidth}{!}{
    \begin{tabular}{lccccc}
        \toprule
        Variant
        & VQAv2
        & MMStar
        & MathVision
        & LogicVista
        & Avg. \\
        \midrule
        Vanilla KD
        & 84.59 & 67.39 & 44.67 & 46.97 & 60.91 \\
        Full MCD
        & \textbf{87.29} & \textbf{73.01} & \textbf{52.37}
        & \textbf{52.67} & \textbf{66.34} \\
       w/o $\ell_{\mathrm{keep}}$
        & 86.85 & 71.65 & 49.83 & 51.23 & 64.89 \\
        w/o $\ell_{\mathrm{effect}}$
        & 86.52 & 71.26 & 49.25 & 50.42 & 64.36 \\
        w/o Verification
        & 85.87 & 70.49 & 48.02 & 49.49 & 63.47 \\       
        \bottomrule
    \end{tabular}
    }
    \caption{Ablation of the causal objectives and verification procedure.}
    \label{tab:causal_supervision_ablation}
\end{table}

\begin{table}[t]
    \centering
    \resizebox{\columnwidth}{!}{
    \begin{tabular}{lccccc}
        \toprule
        Variant
        & VQAv2
        & MMStar
        & MathVision
        & LogicVista
        & Avg. \\
        \midrule
        Random evidence
        & 84.93 & 68.52 & 47.27 & 48.33 & 62.26 \\
        Attention-based evidence
        & 86.51 & 71.08 & 49.89 & 50.86 & 64.59 \\
        Unrestricted replacement
        & 86.34 & 70.49 & 48.56 & 50.00 & 63.85 \\
        Full MCD
        & \textbf{87.29} & \textbf{73.01} & \textbf{52.37}
        & \textbf{52.67} & \textbf{66.34}  \\
        \bottomrule
    \end{tabular}
    }
    \caption{Ablation of causal evidence discovery.}
    \label{tab:evidence_discovery_ablation}
\end{table}

\subsection{Analysis}
\label{sec:analysis}
\begin{figure}[t]
    \centering
    \includegraphics[width=0.9\columnwidth]
    {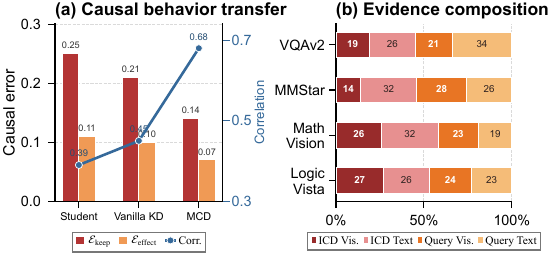}
    \caption{
        Causal behavior transfer and composition of verified evidence.
        (a) Evidence-sufficiency error, causal-response error, and Pearson
        correlation for the original student, Vanilla KD, and MCD.
        Lower errors and higher correlation indicate closer behavior to the
        teacher.
        (b) Composition of verified evidence across four benchmarks.
    }
    \label{fig:causal_analysis}

\end{figure}
\paragraph{Causal behavior transfer.}
We analyze whether MCD transfers the teacher's response to interventions rather
than only its full-prompt prediction. On held-out prompts excluded from evidence screening and training, we apply the same teacher-discovered evidence
and replacements to the original student, Vanilla KD, and MCD. We measure the
evidence-sufficiency error $\mathcal{E}_{\mathrm{keep}}$, causal-response error
$\mathcal{E}_{\mathrm{effect}}$, and Pearson correlation between teacher and
student removal effects, as detailed in
Appendix~\ref{app:causal_metrics}. As shown in
Fig.~\ref{fig:causal_analysis}(a), the original student obtains
$(0.25,0.11,0.39)$ and Vanilla KD improves these values to
$(0.21,0.10,0.45)$. MCD further reaches $(0.14,0.07,0.68)$, reducing
$\mathcal{E}_{\mathrm{keep}}$ and $\mathcal{E}_{\mathrm{effect}}$ by $33.3\%$
and $30.0\%$ relative to Vanilla KD while increasing the correlation by
$0.23$. These results show that MCD more faithfully reproduces the teacher's
response to evidence interventions and its relative strength across examples.

\paragraph{Composition of causal evidence.}
We further analyze how verified evidence is distributed across ICD visual, ICD
text, query visual, and query text tokens on four representative benchmarks.
For each benchmark, we aggregate the fraction of selected tokens from each
source and visualize the resulting distribution as a horizontal percentage
stacked bar in Fig.~\ref{fig:causal_analysis}(b). Query tokens account for
$55\%$ and $54\%$ of the evidence on VQAv2 and MMStar, respectively, whereas ICD tokens constitute $58\%$ on MathVision. LogicVista exhibits a nearly balanced division between ICD and query evidence at $52\%$ and $48\%$.
Across modalities, textual evidence is more prominent on VQAv2 and MMStar
($60\%$ and $58\%$), while MathVision and LogicVista maintain nearly even
visual--textual compositions. These shifts show that the verified evidence
adapts to the task-specific use of the query and ICDs rather than
following a fixed source or modality allocation. We provide an efficiency analysis and a failure case study in Appendix~\ref{app:analysis}.

\section{Conclusion}
This paper introduced MCD, a framework that transfers multimodal ICL capabilities from large-scale LVLMs to smaller LVLMs within the same family. MCD identifies and verifies mechanism-bearing evidence through token-level interventions, then transfers its sufficiency and causal influence to the student. Experiments across three LVLM families and seven benchmarks demonstrate consistent improvements over original student models and existing distillation methods. MCD advances the development of compact LVLMs with stronger multimodal reasoning capabilities.
\section*{Limitations}
MCD is currently applied within a single LVLM family, where the teacher and student share the same tokenizer and image processor. This assumption ensures that the token-level interventions used for causal evidence discovery and verification refer to comparable input units in both models, so that the retained and removed evidence carries a consistent meaning throughout the distillation pipeline. Although same-family distillation is a common setting in prior LVLM distillation work, extending MCD to cross-family teachers and students, in which tokenization schemes and visual encoders differ, would substantially broaden the range of teacher-student pairs to which our framework applies and represents an important future direction.

A second direction concerns the scope of multimodal scenarios considered in this work. Our experiments focus on standard few-shot ICL prompts with a moderate number of ICDs and a single question-answering query. Extending MCD to more general and complex multi-image regimes, including long-context ICL with many interleaved images, tool-augmented multimodal agents, and settings that require interleaved image-text chain-of-thought (CoT) reasoning, is a natural next step. Such extensions would allow the causal supervision provided by MCD to support emerging paradigms such as think-with-image reasoning, where the model must iteratively inspect, reason about, and integrate multiple visual pieces of evidence.
\bibliography{custom}

\appendix

\section{Experimental Setups}
\label{sec:appendix}
\subsection{Training Details}
We train each model for three epochs using the AdamW optimizer with a learning rate of $2\times10^{-5}$, a batch size of 32, cosine decay, and linear warmup. The maximum sequence length is set to 4,096. All experiments are conducted on eight NVIDIA H200 GPUs. The training data are detailed below.
\paragraph{VL-ICL.} VL-ICL is a benchmark specifically designed to evaluate the multimodal ICL capability of LVLMs. Unlike conventional multimodal benchmarks that mainly probe single-image understanding, its few-shot instances are constructed so that the query cannot be answered without first extracting task information from the accompanying ICDs. It covers a diverse set of subtasks including fine-grained visual concept induction, image-to-text rule discovery, visual-textual binding, and interactive multi-image reasoning, forcing models to infer the underlying task specification, output format, and input-output mapping from the ICDs rather than from the query alone.
\paragraph{TrueMICL.} TrueMICL is a benchmark that isolates genuine multimodal ICL from spurious shortcut learning. It rewrites conventional few-shot tasks so that language priors, prompt formatting, and ICD ordering are insufficient to answer the query, and correct predictions require jointly grounding visual content in the ICDs and the query. The benchmark spans multiple task types with tightly controlled ICD-query pairs, enabling a faithful measurement of how well an LVLM integrates cross-modal evidence. Thus, its instances serve as high-quality supervision for training procedures that target evidence-driven multimodal ICL behavior.
\paragraph{SMMILE.} SMMILE is an expert-driven benchmark for multimodal medical in-context learning. It contains few-shot cases curated by clinical experts across a range of medical imaging modalities and diagnostic tasks, with each query paired with ICDs that convey the intended reasoning pattern rather than lexical shortcuts. The benchmark emphasizes fine-grained visual features, domain-specific terminology, and multi-step clinical reasoning, all of which cannot be inferred from the query image alone. As a domain-specific complement to general-purpose ICL benchmarks, SMMILE probes whether LVLMs can extract and reuse specialized visual evidence provided in ICDs.
\paragraph{HatefulMemes.}HatefulMemes is a multimodal classification benchmark for detecting hateful content in internet memes. Each sample pairs an image with an overlaid caption, and the label depends jointly on both modalities: many instances are "benign confounders" whose text or image alone is innocuous but whose combination becomes hateful. This construction penalizes unimodal shortcuts and forces models to reason about the interaction between visual and textual cues.
\paragraph{MME-RealWorld.}MME-RealWorld is a large-scale multimodal benchmark composed of high-resolution real-world images and expert-annotated questions across a broad range of practical scenarios, such as autonomous driving, remote sensing, monitoring, and document understanding. Each question is carefully written to require detailed visual perception and fine-grained reasoning, and the answers cannot be inferred from language priors alone. The benchmark stresses realistic image conditions, spatial precision, and domain-specific knowledge simultaneously.
\paragraph{BlindTest.}BlindTest is a diagnostic benchmark that exposes systematic failures of LVLMs on tasks that would be trivial for a sighted human, such as counting overlapping shapes, identifying intersections between lines, or tracing simple paths. Each item consists of a synthetic image and a short question targeting a single, precisely defined visual property. The benchmark deliberately avoids linguistic complexity so that any error can be attributed to visual perception rather than reasoning.
\paragraph{VisuLogic.}VisuLogic is a benchmark that evaluates visual logical reasoning through structured diagram-based problems, including quantitative reasoning, spatial reasoning, positional reasoning, attribute reasoning, and stylistic reasoning. Each item presents a set of figures arranged as an analogy or a rule-completion task, and answering it requires inferring an abstract transformation from the visual context alone. Because textual descriptions are minimal, the benchmark is largely immune to language-prior shortcuts.
\paragraph{GQA.}GQA is a large-scale visual question answering benchmark built on scene graphs derived from real-world images. Its questions are generated from compositional templates that explicitly target object recognition, attribute prediction, spatial relations, and multi-step reasoning, and each question is paired with a functional program specifying the reasoning steps required. This design yields controlled difficulty and reduced language bias compared with earlier VQA datasets, and it enables fine-grained analysis of model behavior across reasoning types.
\subsection{Benchmarks}
In this section, we introduce the seven multimodal benchmarks used in our experiments. For each benchmark, we adopt its official accuracy metric.
\paragraph{VQAv2.}VQAv2 is a large-scale visual question answering benchmark that pairs natural images with open-ended questions covering object recognition, attribute prediction, counting, and commonsense reasoning. Each question is balanced with a complementary image that produces a different answer, mitigating language priors and forcing models to attend to visual content. Its scale, coverage, and adversarial balancing make it a standard benchmark for evaluating general visual-language competence. In the multimodal ICL setting, VQAv2 is widely adopted to test whether models can leverage a small set of ICDs to align their answers with the expected format and reasoning pattern.
\paragraph{VizWiz.} VizWiz is a visual question answering benchmark collected from blind users, where each question is spoken about a photograph taken by the user. The resulting images often contain motion blur, poor framing, or occlusion, and questions may be ambiguous or even unanswerable. This distribution shift makes VizWiz substantially more challenging than curated benchmarks and demanding of robust cross-modal grounding. It is commonly used to assess whether LVLMs can generalize beyond clean web imagery, and, in ICL evaluation, whether ICDs help models handle noisy inputs and abstain when necessary.
\paragraph{MMStar.} MMStar is a vision-indispensable multimodal benchmark carefully filtered to remove samples that can be answered from text alone. Each question spans six core capabilities, including coarse and fine-grained perception, instance reasoning, logical reasoning, science and technology, and mathematics, ensuring that predictions depend on the accompanying image. The samples are hand-verified to avoid data leakage from common pretraining corpora. Its emphasis on genuine visual grounding makes MMStar a strong probe of multimodal reasoning, and it is frequently used in recent evaluations of state-of-the-art LVLMs to distinguish visually competent models from those relying on language priors.
\paragraph{MathVision.} MathVision is a benchmark for evaluating mathematical reasoning grounded in images, covering geometry, algebra, combinatorics, and other topics drawn from competition-level problems. Each instance consists of a diagram together with a formal problem statement, requiring the model to interpret visual structure and combine it with symbolic manipulation. The benchmark spans multiple difficulty levels and problem types, providing a stringent test of visual-mathematical reasoning. It is widely used to evaluate the higher-order reasoning ability of LVLMs, particularly whether models can move beyond surface recognition to perform structured, multi-step derivations conditioned on visual input.
\paragraph{MDK12.} MDK12 is a multi-discipline benchmark for evaluating reasoning in multimodal large language models across K-12 subjects. Its questions are drawn from real curricula and standardized examinations, and each item combines textual descriptions with diagrams, figures, or tables that are essential to the answer. The benchmark emphasizes long-tail domain knowledge and structured reasoning rather than perceptual pattern matching. Its breadth across disciplines makes it a demanding testbed for measuring whether LVLMs possess the integrated visual, linguistic, and knowledge-based reasoning skills expected of general-purpose multimodal systems.
\paragraph{MMIQ.} MMIQ is a multimodal intelligence-quotient benchmark that adapts classical IQ-test-style problems into vision-language tasks. Items include visual analogies, pattern completion, spatial reasoning, and abstract rule induction, each requiring the model to infer latent transformations from a small set of visual elements. Because the questions rely on abstract, non-verbal patterns rather than world knowledge, MMIQ is largely resistant to shortcuts based on textual priors. It is frequently used to evaluate the abstract and analogical reasoning capabilities of LVLMs, complementing benchmarks that emphasize perception or factual recall.
\paragraph{LogicVista.} LogicVista is a benchmark for evaluating logical reasoning grounded in visual inputs. It covers deductive, inductive, and abductive reasoning across categories such as syllogisms, propositional logic, spatial logic, and diagrammatic reasoning, with each instance providing a visual scene or figure together with a logical question. Answers cannot be obtained from either the image or the text in isolation, requiring genuine multimodal integration. Its focused design targets a capability that is often underrepresented in general multimodal benchmarks, making it a useful probe of whether LVLMs can perform structured logical inference over visual evidence.
\subsection{Baselines}
In this section, we introduce the three baseline methods used in our experiments. For methods that perform distillation in multiple stages, we apply only distillation fine-tuning (DFT) and omit subsequent supervised fine-tuning (SFT) to ensure a fair comparison.
\paragraph{LLaVA-KD.} LLaVA-KD extends standard output distillation with relational distillation tailored to LVLMs. In addition to matching the teacher's output distribution, it aligns pairwise relations between hidden states across samples and modalities, encouraging the student to reproduce the teacher's overall representation geometry. This relational objective is intended to capture cross-modal structure that pointwise alignment omits.
\paragraph{CompoDistill.} CompoDistill focuses on improving the compositional reasoning capability of LVLMs by distilling intermediate attention distributions from the teacher to the student. It aligns attention maps over image and text tokens at selected layers, providing finer-grained supervision than output-only distillation. This design encourages the student to reproduce the teacher's token-interaction pattern rather than merely its final predictions. CompoDistill is included as a representative attention-based distillation baseline.
\paragraph{Align-TI.} Align-TI is a recent LVLM distillation method that combines cross-modal alignment and attention distribution matching to capture diverse token interactions. It transfers alignment between textual queries and visual tokens while also aligning the teacher's attention patterns across selected layers, aiming to reproduce both the semantic and structural aspects of the teacher's forward pass. It represents the current state of the art among attention-based distillation approaches for LVLMs and, in this work, is adapted to multi-image ICL prompts.
\section{Attribute-matched Replacement}
\label{app:replacement}
For each prompt $\boldsymbol{x}$, we sample a different training example $\boldsymbol{x}^{0}$ as the source of replacement content. The replacement example is selected once during preprocessing using a deterministic seed derived from the identifier of $\boldsymbol{x}$ and is reused throughout training. For a text token $u_j$, we define its attributes by its prompt role, textual field, and within-field position. The prompt role indicates whether the token belongs to an ICD or the query, while the textual field distinguishes an ICD question, an ICD answer, and the query question. If the token is the $t$-th token of a field containing $L$ tokens, its position is normalized as $p_j=(t+0.5)/L$. We select $u_j^0$ from the same role and textual field in $\boldsymbol{x}^{0}$ with the closest normalized position. When several tokens are equally close, one is selected deterministically, and an identical token is skipped whenever another valid replacement is available. For a projected visual token, the attributes consist of its ICD or query role and its normalized spatial coordinates. Specifically, a token at row $h$ and column $w$ of an $H\times W$ visual-token grid is assigned coordinates $((h+0.5)/H,(w+0.5)/W)$. Its replacement is taken from the corresponding image role in $\boldsymbol{x}^{0}$ at the nearest normalized grid location, which reduces to the same row and column when the two grids have identical sizes. For ICD content, the replacement ICD in $\boldsymbol{x}^{0}$ is sampled uniformly from its ICDs and then used consistently for all tokens belonging to the same original ICD. If the selected example does not contain a valid token with the required role or field, another example is sampled instead of relaxing the matching constraints. The replacement changes only the input embedding at the original token position, leaving sequence length, structural tokens, image boundaries, positional indices, and the attention mask unchanged. For teacher and student models, the same replacement token identity or replacement image location is used, but its embedding $b_{M,j}$ is independently produced by the embedding and visual projection modules of model $M\in\{T,S\}$.
\section{Additional Ablation Study}
\subsection{Sensitivity to Hyperparameters}
\label{app:hyperparameter_sensitivity}
\begin{figure*}[t] \centering \includegraphics[width=\textwidth]{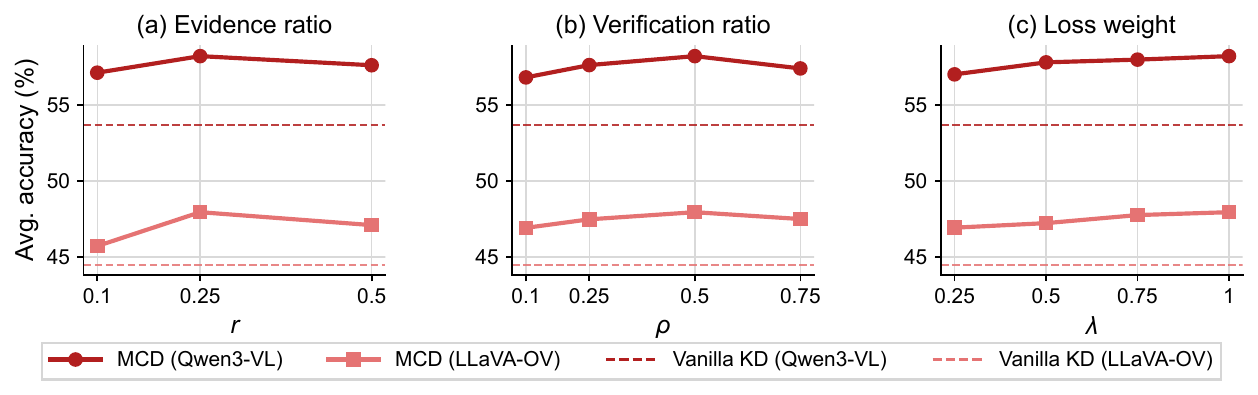} \caption{Sensitivity of MCD to three hyperparameters: evidence retention ratio $r$, verification threshold $\rho$, and causal loss weight $\lambda$. Results are averaged across the seven evaluation benchmarks. Dashed lines denote Vanilla KD as reference.} \label{fig:app1} \end{figure*}

We study the three hyperparameters that control the coverage, reliability, and
strength of causal supervision. The evidence ratio $r$ determines the fraction
of candidate tokens selected as evidence, the verification ratio $\rho$
determines the fraction of highest-scoring prompts retained after causal-effect
verification, and $\lambda$ weights the MCD loss relative to conventional
supervision. We vary one hyperparameter at a time while fixing the others to
their default values, $r=0.25$, $\rho=0.50$, and $\lambda=1.0$. Fig.~\ref{fig:app1} reports
Qwen3-VL and LLaVA-OneVision separately and the default points correspond to their main results of $58.22$ and $47.95$.

For the evidence ratio, increasing $r$ from $0.10$ to $0.25$ improves
Qwen3-VL from $57.13$ to $58.22$ and LLaVA-OneVision from $45.72$ to $47.95$.
Increasing it further to $0.50$ reduces the scores to $57.62$ and $47.10$.
Selecting too few tokens can omit complementary query and ICD information,
an effect that is particularly pronounced for LLaVA-OneVision, whereas
selecting too many introduces weakly relevant content and reduces the contrast
between the retain-evidence and remove-evidence prompts. The verification
sweep follows a similar single-peak pattern. Moving $\rho$ from $0.10$ to
$0.50$ raises the two scores from $56.82$ and $46.92$ to $58.22$ and $47.95$,
while retaining $75\%$ of prompts lowers them to $57.41$ and $47.50$. This
result reflects the trade-off between retaining sufficient causal supervision
and excluding prompts with weak intervention effects. Finally, increasing
$\lambda$ from $0.25$ to $1.0$ steadily improves Qwen3-VL from $57.02$ to
$58.22$ and LLaVA-OneVision from $46.93$ to $47.95$. Both model families exhibit consistent trends, supporting the default configuration as a balanced choice for evidence coverage, verification reliability, and objective strength. MCD also remains robust across different hyperparameter settings.

\subsection{Scope of Causal Supervision}
\label{app:supervision_scope}
We next determine whether the improvements arise from one dominant modality
or prompt component. We restrict evidence discovery and intervention to text
tokens, visual tokens, ICD tokens, or query tokens, producing Text-only,
Vision-only, ICD-only, and Query-only variants. Each variant uses the same
selection ratio within its eligible candidate set, while the original prompt
and standard distillation objective remain unchanged. Table~\ref{tab:scope_ablation}
shows that Text-only and Vision-only obtain average scores of 64.95 and 64.24,
respectively, while ICD-only and Query-only obtain 64.50 and 64.87. The stronger
Text-only result reflects the importance of questions and ICD
answers for identifying the task mapping, while the remaining improvement of
Vision-only confirms that the causal signal is not reducible to language patterns. Query-only performs better than ICD-only because the query directly
specifies the prediction target, yet both remain below full MCD at 66.34. The
combined result therefore indicates that effective multimodal ICL depends on
jointly transferring visual and textual evidence from both the ICDs
and the query, matching the causal pathway that motivates MCD.

\begin{table}[t]
    \centering
    \resizebox{\columnwidth}{!}{
    \begin{tabular}{lccccc}
        \toprule
        Variant
        & VQAv2
        & MMStar
        & MathVision
        & LogicVista
        & Avg. \\
        \midrule
        Text-only
        & 86.79 & 70.62 & 51.18 & 51.20 & 64.95 \\
        
        Vision-only
        & 85.72 & 69.85 & 50.83 & 50.54 & 64.24 \\
        
        ICD-only
        & 85.94 & 70.50 & 50.79 & 50.76 & 64.50 \\
        
        Query-only
        & 86.45 & 70.28 & 51.31 & 51.42 & 64.87 \\
        Full MCD
        & \textbf{87.29} & \textbf{73.01} & \textbf{52.37}
        & \textbf{52.67} & \textbf{66.34} \\
        \bottomrule
    \end{tabular}
    }
    \caption{Ablation of the modality and ICL prompt component receiving causal supervision.}
    \label{tab:scope_ablation}
\end{table}

\subsection{Decomposition of Evidence Verification}
\label{app:verification_decomposition}
The main experiments establish the benefit of verification as a whole. We
further examine whether its retain-evidence and remove-evidence measurements
provide complementary information. To control the amount of causal training
data, all verified variants accept the same proportion $\rho$ of examples.
Retain-only ranks examples by the similarity between the full and
retain-evidence predictions, Remove-only ranks them by the divergence caused
by evidence removal, and Joint verification uses their difference as defined
in the main method. Table~\ref{tab:verification_decomposition} shows that both
one-sided criteria improve upon the unverified setting, with Remove-only
reaching 65.16 and Retain-only reaching 64.73. Neither criterion alone matches
the 66.34 achieved by Joint verification. Retention alone may select evidence
that reproduces the answer but is behaviorally redundant, while removal alone
may favor destructive perturbations that do not preserve sufficient task
information. Their combination requires the same evidence to be sufficient
when retained and influential when removed, providing a stricter operational test of the teacher's causal dependence and explaining the benefit of the two-sided verification design.

\begin{table}[t]
    \centering
    \resizebox{\columnwidth}{!}{
    \begin{tabular}{lccccc}
        \toprule
        Variant
        & VQAv2
        & MMStar
        & MathVision
        & LogicVista
        & Avg. \\
        \midrule
        No verification
        & 85.87 & 70.49 & 48.02 & 49.49 & 63.47 \\
        Retain-only
        & 86.35 & 71.33 & 49.58 & 51.64 & 64.73 \\
        Remove-only
        & 86.51 & 71.67 & 50.25 & 52.20 & 65.16 \\
        Joint verification
        & \textbf{87.29} & \textbf{73.01} & \textbf{52.37}
        & \textbf{52.67} & \textbf{66.34} \\
        \bottomrule
    \end{tabular}
    }
    \caption{Decomposition of the evidence verification criterion. All
    verified variants retain the same proportion of training examples.}
    \label{tab:verification_decomposition}
\end{table}

\subsection{Robustness to Replacement Sampling}
\label{app:replacement_stability}
Because MCD constructs interventions using a randomly selected matched
replacement example, we test whether its result depends on one favorable
replacement assignment. For Fixed-$K$, we independently sample $K$ matched
replacement examples during preprocessing and average their teacher
attributions and intervention responses before selecting and verifying the
evidence. We additionally consider epoch-wise resampling, which draws one new
replacement for every training epoch and recomputes the corresponding teacher
metadata. Table~\ref{tab:replacement_stability} reports the average score
under each setting. Increasing $K$ from one to four changes the average score
by only 0.19, while teacher preprocessing grows approximately linearly because
each additional replacement requires another set of attribution and
intervention computations. The changes on individual benchmarks are also
small, with Fixed-4 improving over Fixed-1 by at most 0.53 points. In contrast,
epoch-wise resampling reaches 65.42, which is 0.92 points below the fixed
default, with larger drops on MathVision and LogicVista. Thus, changing the causal
target across epochs particularly disrupts tasks that require
multi-step use of the ICDs. These results show that MCD is
insensitive to the fixed replacement assignment and support
Fixed-1 as the best balance between performance and preprocessing cost.

\begin{table}[t]
    \centering
    \resizebox{\columnwidth}{!}{
    \begin{tabular}{lccccc}
        \toprule
        Sampling strategy
        & VQAv2
        & MMStar
        & MathVision
        & LogicVista
        & Avg. \\
        \midrule
        Fixed-1 (default)
        & 87.29 & 73.01 & 52.37 & 52.67 & 66.34 \\
        Fixed-2
        & 87.14 & \textbf{73.13} & 52.31 & 53.12 & 66.43 \\
        Fixed-4
        & \textbf{87.42} & 72.79 & \textbf{52.72}
        & \textbf{53.20} & \textbf{66.53} \\
        Epoch-wise resampling
        & 86.69 & 72.31 & 51.22 & 51.45 & 65.42 \\
        \bottomrule
    \end{tabular}
    }
    \caption{Robustness to replacement sampling. Fixed-$K$ averages teacher
    signals from $K$ independently sampled matched replacements.}
    \label{tab:replacement_stability}
\end{table}
\section{Additional Analysis}
\label{app:analysis}
\subsection{Metrics for Causal Behavior Transfer}
\label{app:causal_metrics}

Let $\mathcal{H}$ denote the held-out prompt set. For each
$\boldsymbol{x}\in\mathcal{H}$, we use the evidence set $E_{\boldsymbol{x}}$
discovered by the teacher to construct the same full, retain-evidence, and
remove-evidence prompts for every evaluated student. All models are evaluated
on the fixed teacher-generated answer
$\widetilde{\boldsymbol{y}}_T$ and its prefixes. Let $L_{\boldsymbol{x}}$ be
its length and let $\bar P_{M,k}^{v}$ denote the distribution of model $M$ for
prompt variant $v\in\{F,K,D\}$ using the sparse support and tail aggregation
defined in Section~\ref{sec:causal_distillation}. For an evaluated student $M$, we compute

{\small
\begin{align}
    e_{\mathrm{keep}}^{M}(\boldsymbol{x})
    &=
    \frac{1}{2L_{\boldsymbol{x}}}
    \sum_{k=1}^{L_{\boldsymbol{x}}}
    \left\|
        \bar P_{T,k}^{F}-\bar P_{M,k}^{K}
    \right\|_1,
    \label{eq:analysis_keep_metric}\\
    e_{\mathrm{effect}}^{M}(\boldsymbol{x})
    &=
    \frac{1}{4L_{\boldsymbol{x}}}
    \sum_{k=1}^{L_{\boldsymbol{x}}}
    \left\|
        (\bar P_{T,k}^{F}-\bar P_{T,k}^{D})
        -(\bar P_{M,k}^{F}-\bar P_{M,k}^{D})
    \right\|_1
    \label{eq:analysis_effect_metric}
\end{align}
}
The reported errors average these quantities over the held-out set,
$\mathcal{E}_{\mathrm{keep}}^{M}
=|\mathcal{H}|^{-1}\sum_{\boldsymbol{x}}e_{\mathrm{keep}}^{M}(\boldsymbol{x})$
and
$\mathcal{E}_{\mathrm{effect}}^{M}
=|\mathcal{H}|^{-1}\sum_{\boldsymbol{x}}e_{\mathrm{effect}}^{M}(\boldsymbol{x})$.
To measure whether the student preserves variation in the strength of the teacher's intervention response, we define the removal effect
\begin{equation}
    r_M(\boldsymbol{x})
    =
    \frac{1}{L_{\boldsymbol{x}}}
    \sum_{k=1}^{L_{\boldsymbol{x}}}
    D_{\mathrm{JS}}
    \left(
        \bar P_{M,k}^{F}
        \middle\|
        \bar P_{M,k}^{D}
    \right)
    \label{eq:analysis_removal_effect}
\end{equation}
and report the Pearson correlation
{\small
\begin{align}
    \operatorname{Corr}(T,M)
    &=
    \frac{1}{Z_TZ_M}
    \sum_{\boldsymbol{x}\in\mathcal{H}}
    \bigl(r_T(\boldsymbol{x})-\mu_T\bigr)
    \bigl(r_M(\boldsymbol{x})-\mu_M\bigr),
    \label{eq:analysis_correlation}
    \\
    Z_J
    &=
    \left[
        \sum_{\boldsymbol{x}\in\mathcal{H}}
        \bigl(r_J(\boldsymbol{x})-\mu_J\bigr)^2
    \right]^{1/2},J\in\{T,M\}.
    \label{eq:analysis_correlation_scale}
\end{align}
}
\subsection{Efficiency Analysis}
\label{app:efficiency_analysis}
\begin{table*}[h]
    \centering
    \resizebox{0.92\textwidth}{!}{
    \begin{tabular}{lcccccc}
        \toprule
        Method
        & Teacher processing
        & Teacher in training
        & Student evals./epoch
        & Student evals./3 epochs
        & Cache (GiB)
        & Inference cost \\
        \midrule
        Vanilla KD
        & $F_T$ once & No & $1.00F_S$ & $3.00F_S$ & $0.34$ & $1.00\times$ \\
        LLaVA-KD
        & $F_T$/epoch & Yes & $1.00F_S$ & $3.00F_S$ & -- & $1.00\times$ \\
        Align-TI
        & $F_T$/epoch & Yes & $2.00F_S$ & $6.00F_S$ & -- & $1.00\times$ \\
        MCD
        & $4F_T+B_T^{\mathrm{in}}$ once & No & $1.75F_S$ & $5.25F_S$
        & $0.80$\textsuperscript{$\dagger$} & $1.00\times$ \\
        \bottomrule
    \end{tabular}
    }
    \\[2pt]
    \parbox{0.92\textwidth}{\footnotesize
    $\dagger$ Excludes the compact evidence mask, whose exact size is
    $\sum_{i=1}^{D}\lceil N_i/8\rceil$ bytes. A dash indicates that the original online-teacher
    implementation does not use an offline teacher cache.}
    \caption{Algorithmic efficiency on the fixed 60K-example training set.
    Counts are per example; the three-epoch column includes only student
    evaluations. Cache values use $K_{\mathrm{cache}}=128$, $\bar L=8$, and
    $\bar a=0.5$. Inference cost is normalized by the original student.}
    \label{tab:efficiency}
\end{table*}

\begin{table*}[h]
    \centering
    \small
    \begin{tabular}{
        >{\raggedright\arraybackslash}p{2.4cm}
        >{\raggedright\arraybackslash}p{4.2cm}
        >{\raggedright\arraybackslash}p{4.6cm}
        >{\raggedright\arraybackslash}p{3.5cm}}
        \toprule
        Case type
        & Concrete multimodal ICL scenario
        & Why the case is challenging
        & Potential extension \\
        \midrule
        Coherence-sensitive intervention
        & OCR strings, diagram labels, or relational statements must remain
          consistent with nearby visual content.
        & A structurally matched replacement creates an implausible local
          image-text pair and induces a response unrelated to the target
          mechanism.
        & Use context-conditioned replacements that preserve local semantic
          consistency. \\
        \addlinespace
        Teacher uncertainty
        & Low-quality images, rare mechanisms, or open-ended questions admit
          uncertain or multiple valid predictions.
        & Evidence rankings and removal effects change across valid teacher
          responses, or the prompt is removed by correctness screening.
        & Use multi-response or ensemble verification to marginalize over
          acceptable teacher predictions. \\
        \bottomrule
    \end{tabular}
    \caption{Representative challenging cases and potential extensions of
    MCD in multimodal ICL.}
    \label{tab:failure_cases}
\end{table*}

Here, we report model evaluations because
the three model families use different decoder kernels and model-parallel
layouts. This gives a reproducible comparison that is independent of device
utilization. Let $F_T$ denote one teacher-forced evaluation of a complete
prompt--answer sequence, $B_T^{\mathrm{in}}$ the backward pass required only
for the gradient with respect to the input embeddings, and $F_S$ one student
evaluation. On the fixed 60K-example training set, Vanilla KD preprocesses one
$F_T$ per example. MCD additionally evaluates the interpolated prompt, its
input gradient, and the retain- and remove-evidence prompts. Its one-time
teacher cost is therefore $4F_T+B_T^{\mathrm{in}}$ per example when the
full-prompt distribution produced during correctness screening is reused. The
teacher is then removed from training. At each epoch, the conventional loss
uses one $F_S$. An accepted example uses one additional $F_S$ for the keep
objective or two for the effect objective with equal probability. Hence, the
expected student cost is
\begin{equation}
    1+\bar a\!\left[\tfrac{1}{2}(1)+\tfrac{1}{2}(2)\right]
    =1+1.5\bar a
    \label{eq:student_efficiency}
\end{equation}
evaluations per example and epoch. With the realized acceptance rate
$\bar a=0.5$, MCD uses $1.75F_S$ per epoch, or $5.25F_S$ over three epochs,
compared with $3F_S$ for Vanilla KD. The prompt variants are evaluated
sequentially, so this increase affects total computation but does not multiply
peak activation memory.

We further calculate cache storage from the representation used by the loss.
With $K=K_{\mathrm{cache}}$, a Vanilla-KD distribution stores $K$ uint32 token
indices, $K$ BF16 probabilities, and one BF16 tail mass, requiring $6K+2$
bytes per answer token. For an accepted MCD example, the union of the full- and
remove-prompt supports contains at most $2K$ indices, and two BF16
distributions are stored on this union. This adds at most $16K+4$ bytes per
answer token. For $D$ examples with mean answer length $\bar L$, the total
distribution cache is therefore
\begin{equation}
    C_{\mathrm{cache}}
    \leq
    D\bar L\big[(6K+2)+\bar a(16K+4)\big]\ \text{bytes}.
    \label{eq:cache_size}
\end{equation}
For $D=60{,}000$, $K=128$, $\bar a=0.5$, and $\bar L=8$, this is $0.80$~GiB;
the bit-packed evidence masks add
$\sum_{i=1}^{D}\lceil N_i/8\rceil$ bytes. Table~\ref{tab:efficiency}
summarizes the
result. MCD has a larger one-time teacher cost than output KD, but avoids a
teacher in the training loop and remains lighter in student evaluations than
the two-evaluation token-interaction objective. All methods retain the
unmodified student architecture and consequently have identical deployment
cost.

\subsection{Failure Case Study}
\label{app:failure_cases}
To delineate the current scope of MCD, we examine representative prompts that
remain challenging after distillation and summarize two recurring scenarios
in Table~\ref{tab:failure_cases}. First, structural matching does not guarantee
semantic plausibility. Replacing an OCR string, diagram label, or relational
statement with a position-matched token from another example can break local
image-text consistency, so part of the output change may reflect an
off-manifold prompt rather than removal of the intended mechanism evidence.
The verification procedure filters many such cases, while a replacement model
conditioned on local visual and textual context could provide more precise
interventions. Second, MCD inherits uncertainty from the teacher. For
low-quality VizWiz images, rare task mechanisms, or open-ended questions with
multiple valid answers, the selected evidence can vary with the teacher
response even when several predictions are acceptable. Correctness and
causal-effect screening already reduce this noise, while multi-response or
ensemble verification could further preserve useful supervision when several
teacher predictions are reasonable. These cases do not affect the inference
cost or general applicability of MCD, but identify two promising directions
for making its causal supervision more context-aware and uncertainty-aware.

\end{document}